# Artificial intelligence for Sustainability in Energy Industry: A Contextual Topic Modeling and Content Analys

Tahereh Saheb[1]
Research Assistant Professor,
Science & Technology Studies Group,
Management Studies Center,
Tarbiat Modares University, Tehran, Iran
t.saheb@modares.ac.ir

Mohammad Dehghani
Industrial and Systems Engineering
Tarbiat Modares University
Tehran, Iran
mohamad.dehqani@modares.ac.ir

*Abstract—* **Parallel to the rising debates over sustainable energy and artificial intelligence solutions, the world is currently discussing the ethics of artificial intelligence and its possible negative effects on society and the environment. In these arguments, sustainable AI is proposed, which aims at advancing the pathway toward sustainability, such as sustainable energy. In this paper, we offered a novel contextual topic modeling combining LDA, BERT and Clustering. We then combined these computational analyses with content analysis of related scientific publications to identify the main scholarly topics, sub-themes and cross-topic themes within scientific research on sustainable AI in energy. Our research identified eight dominant topics including sustainable buildings, AI-based DSSs for urban water management, climate artificial intelligence, Agriculture 4, convergence of AI with IoT, AI-based evaluation of renewable technologies, smart campus and engineering education and AI-based optimization. We then recommended 14 potential future research strands based on the observed theoretical gaps. Theoretically, this analysis contributes to the existing literature on sustainable AI and sustainable energy, and practically, it intends to act as a general guide for energy engineers and scientists, AI scientists, and social scientists to widen their knowledge of sustainability in AI and energy convergence research.**

---

[1] Corresponding Author

1. **Introduction**

The rise of unsustainable practices and procedures co-occurred with the rising urbanization and civilization have driven the emergence of AI- based solutions to assist the path toward sustainability [1–3]. Excessive consumption and unsustainable energy sources, which have increased at an unprecedented rate due to factors such as urbanization, improper building construction, transportation, environmental changes, and population growth, have pressured the energy industry to pursue clean energy sources and smart solutions [4]. The deployment of alternative energy sources and access to sustainable energy are pillars of global economic growth [5] and fight against environmental hazards, in particular climate change [6]. Thus, the energy sector has focused its efforts not only on developing new sources of energy, but also on inventing novel technical solutions that increase the efficiency of existing mitigation measures [7]. AI-based interventions, which are available in the form of both hard and soft solutions, such as robots and algorithms and models, are one of these solutions that have come to assist humanity [8]. Artificial intelligence can provide a wide range of intelligent solutions, from predictive and prescriptive energy consumption insights to intelligent energy generation and distribution.

Parallel to the escalating discussions over sustainable energy and artificial intelligence solutions, the world is now debating the ethics of artificial intelligence and its potentially negative effects on society and the environment. Ethical AI considers not just AI's moral dimensions, but also its epistemic perspectives [9]. While prior studies have urged scholars to focus on the epistemological aspects of sustainable AI and to open the black box of algorithms to develop sustainable models and algorithms [10], other researches have concentrated on AI for social good and its favorable societal and environmental circumstances [11,12]; such as the development of sustainable AI.

In this article, we define sustainable AI as AI that is designed to achieve sustainability and is called AI for sustainability, as differed from AI that is designed to be sustainable and is called sustainability of AI [10]. In this paper, the term "sustainable AI" refers to the extent to which artificial intelligence can help society accomplish their sustainability goals [13,14]. The energy industry is one of the core industries that will benefit from sustainable AI, which will aid in the development of energy sustainability [15]. Sustainable energy strives to fulfill today's energy demand without depleting energy supplies or harming the environment. Sustainable energy systems are regarded as a requirement for achieving all the Sustainable Development Goals (SDGs) [16]. Sustainable artificial intelligence can help to expedite the development of sustainable energy [14]. To advance sustainable energy, the industry has supplied a wide variety of choices, including wind energy, fossil fuels, solar energy, and bioenergy. It's also vital to recognize how academics have dealt with the confluence of sustainability, artificial intelligence, and energy.

This research is novel from various perspectives. First, this study intends to foster discussions on sustainable AI by identifying the most important research issues in the area, highlighting intellectual gaps, and proposing potential research streams. It is obvious that the energy sector and scientific research and innovation are inextricably linked. Scientific research is seen to be the cornerstone of technological advancements [17]. Identifying the intellectual frameworks of scientific research across time and the historical progression of its themes can have a huge influence on the effectiveness or failure of new technological solutions. To our knowledge, scientific research on sustainable energy is lacking a coherent understanding of how artificial intelligence has been integrated into this domain and how it should be conducted in the future. It is therefore imperative to perform a mixed-method literature review to have a deeper understanding of the deployment of AI to achieve sustainable energy in order to identify existing research gaps and potential future research streams. The second aspect of this research that distinguishes it from prior research is its novel methodology. Extensive literature reviews are conducted by scholars using bibliometric methodologies [18–20] or topic modeling techniques such as Latent Dirichlet Allocation (LDA) [21,22] or qualitative content analysis [23]. As a result, we incorporated all the aforementioned review methodologies to ensure that their findings were complementary. Furthermore, because both bibliometric and LDA topic modeling are based on keyword co-occurrence analysis, we included a contextual embedding-based topic modeling analysis that incorporates use of sentences as fundamental units of analysis. This method which is the latest development in natural language processing (NLP) is offered by Google under the name of Bidirectional Encoder Representations for Transformers (BERT) [24] . BERT makes use of the Transformer library, which uses machine learning to discover contextual relationships between words in a text. Our integrated adoption of computational and advanced topic modeling tools, as well as qualitative analysis, enables us to gain highly objective, coherent, superior, and meta-analytical insight into present research on sustainable artificial intelligence in energy and to forecast its future. The final contribution of this research is that we offer a thorough list of research gaps and potential research agendas that may be used to increase the depth of research on sustainable artificial intelligence in the energy industry

In sum, the theoretical contribution of this research is to extent the literatures on sustainable AI and sustainable energy by determining the key academic themes, sub-themes and cross-topic common themes addressed by scientists working on sustainable AI in energy, as well as how these subjects have evolved over time.  Practically, this research attempts to enlighten policymakers, the energy sector, and engineers and developers of artificial intelligence about the productivity of science while emphasizing the challenges that require more AI-based responses. Additionally, it encourages policymakers to design artificial intelligence regulations that promote the development of sustainable AI in the energy sector while mitigating the unintended consequences of unsustainable energy sources and AI solutions.

The study is structured as follows: we begin with an explanation of our methodology and then go on to the findings, which include our topic modeling and content analysis of topics. We conclude the study by discussing our findings, theoretical research gaps, and potential future research directions. We also discussed the theoretical and practical contribution of the study. We conclude the paper with a conclusion.

2. Methodology

It is a widely held belief among researchers that each quantitative and qualitative research technique has inherent strengths and weaknesses; hence, combining both methods is advised to ensure that their results complement one another. We drew on and included four complimentary sets of research methodologies in our study. Three of these, BERT, LDA topic modeling and clustering are connected with text mining techniques. Additionally, we supplemented these quantitative findings with a qualitative topic-based content analysis. Our mixed-methods approach is new in three ways. First, we employed computational approaches such as BERT, LDA, and clustering to discover the thematic content of research on sustainable AI in energy. Second, we conducted a comprehensive analysis of the retrieved topics using content analysis as a qualitative approach. Third, we integrated LDA and BERT topic modeling approaches in this study to achieve the highest level of topic identification accuracy. Our suggested mixed-method methodology may be used by researchers from a variety of disciplines to improve our understanding of quantitative and computational analyses through the use of topic-based content analysis.

LDA is predicated on the premise that documents are made of topics and that some words are more likely to occur in certain topics than others (Xie *et al.*, 2020). While LDA has been regularly used by academics to identify topics, it does have some limitations due to the fact that it is a word co-occurrence analysis and so cannot incorporate the entire content of the sentence. Additionally, it does not do well on short texts [26]. Additionally, the outcomes of LDA may be challenging for humans to comprehend and consume [27]. By contrast, BERT topic modeling is focused on detecting semantic similarity and integrating topics with pre-trained contextual representations [28] It substantially enhances the coherence of neural topic models by including contextual information into the topic modeling process [29]. BERT makes use of the Transformer library, which has an Autoencoder technique: an encoder that scans the text input. We combined the LDA and BERT vectors in this study to improve topic recognition and clustering. Moreover, because one of the most difficult aspects of word-sentence embedding is dealing with high dimensions, we applied the Uniform Manifold Approximation and Projection (UMAP) approach. In comparison to other approaches, UMAP is one of the most efficient implementations of manifold learning [30].

*1.2. Corpus Building*

On May 29th 2021, we searched the following keywords inside the title, keyword, and abstract: "artificial intelligence" OR "AI" AND "sustainable" OR "sustainability" AND "energy". This search resulted in the retrieval of 981 documents. Following that, we restricted the document type to Articles and the language to English. This exclusion resulted in 296 articles. Following that, we manually evaluated the titles and abstracts of the articles to identify the most pertinent ones that examined the role of artificial intelligence in ensuring the energy sector's sustainability. This screening yielded 182 publications spanning the years 2004 to 2022. Given that abstracts of research articles are the most succinct summary of key ideas [22], we included abstracts of the final publications in the study's corpus.

*2.2. Preprocessing and Post-Processing Stages*

Python 3.7.9 was utilized for pre- and post-processing, as well as for topic modeling analysis. We preprocessed our corpus using the NLTK and Scikit-learn packages, as well as Regular Expressions or RegEX. We import the word tokenize from the NLTK to begin the tokenization process. After removing punctuation, we lowercased our characters and deleted all numeric characters, punctuation, and whitespace. Additionally, we eliminated no-word repetitions and anything enclosed in parenthesis. Additionally, we eliminated the NLTK library's stopwords.

We reviewed the first findings and created a manual exclusion list for more relevant topic identification during the postprocessing step. We added the core keywords (i.e. artificial intelligence, AI, energy, sustainable, sustainability) in the exclusion list to enhance the coherence of the findings. We used stemming throughout the preprocessing step; however, after observing the first results, we decided to remove the stemming to make the words displayed in the word clouds more understandable. We next used the lemmatization procedure, which we abandoned following the findings of the word clouds in order to make our topic labeling approach more comprehensible. Additionally, we estimated the TF-IDF score for each word in the corpus. We eliminated words with scores that were lower than the median of all TF-IDF values. We calculated the TF-IDF scores using the Scikit-learn package. The maximum TF-IDF score was set to 0.8 and the minimum value at 0.11. Additionally, we incorporated unigrams and bigrams.

*3.2. Topic Modeling*

We applied the following libraries to conduct the topic modeling: Pandas to read the dataset, Gensim to perform LDA, Transformers to perform BERT, Keras to perform auto-encoding, and Seaborn and Matplotlib to visualize the results. We imported the TFID vectorizer from the Scikit-learn feature extraction and KMeans from the Scikit-learn cluster. The probabilistic topic assignment vector was constructed using LDA, while the sentence embedding vector was constructed using BERT. To begin, we used the TF-IDF,

LDA, and BERT to model the topics (Figure 1). The LDA and BERT vectors were then concatenated in order to balance the information content of each vectors. We incorporated the Keras package to process the auto-encoder in order to learn a lower-dimensional latent space representation for the concatenated vector. To ensure the clusters were of good quality, we calculated the Silhouette Score, which was 0.566 and near to one for LDA+BERT+ Clustering. TFIDF+clustering received a score of 0.048, while BERT+clustering received a score of 0.095 (Figure 2). The Silhouette Score is used for cluster quality [31]. The score ranges from -1 to 1. If the score is near to one, the cluster is dense and well isolated from neighboring clusters. In comparison to other topic modeling techniques, LDA BERT Clustering is closer to 1, indicating that the clusters are of excellent quality.

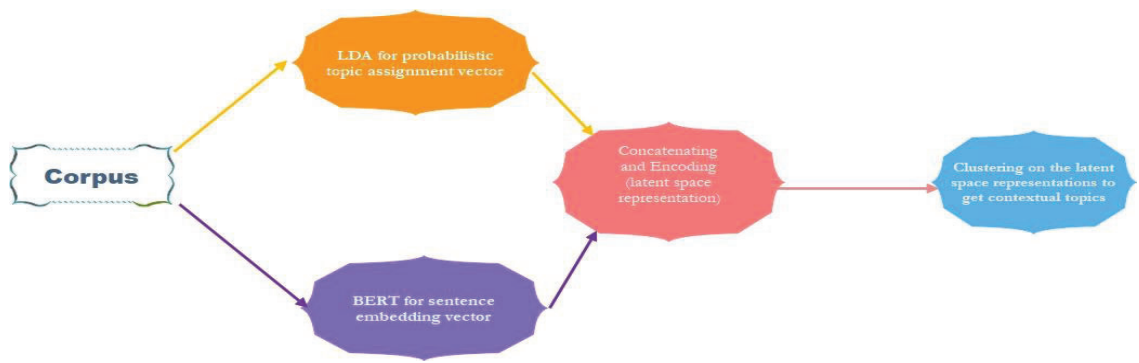

*Figure 1 The concatenating and encoding LDA and BERT vectors to extract contextual topics*

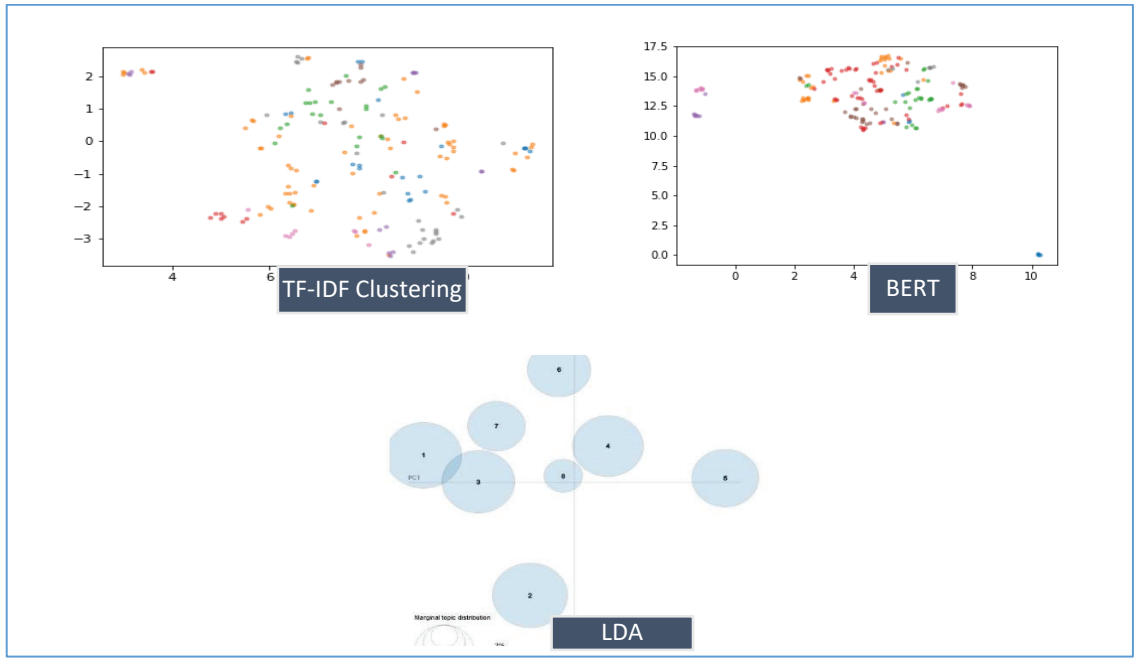

*Figure 2 The separate and independent results of topic modeling of research on sustainable AI in energy by using TF-IDF, BERT and LDA algorithms*

The final topic identification obtained by LDA+BERT+Clustering Algorithms is depicted in Figure 3. We utilized the UMAP package to do dimension reductions and set the topic count to eight. We also evaluated several topic clustering, including 10, 4, and 6. The authors determined that eight topics were better separated from one another and had a greater density within each topic; this demonstrates the excellent quality of clustering. As indicated by the percentage of documents contained inside each topic, approximately 11% of documents belong to topic 0 and approximately 16% to topic 1. Clustering resulted in a balanced distribution of documents within each topic, confirming the clustering's excellent quality.

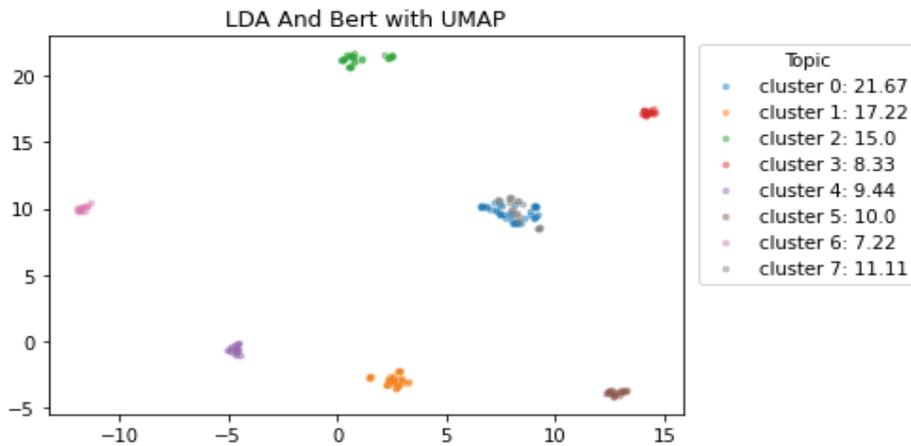

*Figure 3 The global view of the topic model on sustainable AI in energy research area. We integrated LDA, BERT and clusetering for topic modeling detection.*

## 3. Results

*1.3. Descriptive Analysis*

Figure 3.0 shows a representation of the topic model on sustainable AI in energy research field with respect to the overall global view. This visualization represents the topic modeling results, where topics are illustrated as clusters on a two-dimensional plane. Also shown in Figure 4 is the word cloud visualization of the topics with the most frequently used terms in each topic. Topics 1, 2, and 3 represent the greatest research interest in the model based on 8 topics and including 21.67%, 17.22%, and 15.0% of the corpus. Our research uncovered eight different topics. These topics will be described, and then a content analysis of the papers that are associated with each one will be carried out throughout this part of the article.

These articles were organized according to their relative likelihood of belonging to each topic. As seen in Figure 4.0, the three most-covered topics by academia are topic 1: Sustainable buildings (22.5%), Topic 2: AI-based DSSs for urban water management (16.5%) and Topic 3: Climate Artificial Intelligence (14.8%). About 54% of the articles in the corpus are concerned with these three themes.

The word cloud visualization (Figure 6.0) shows the identified topics after labeling based on the topic three keywords. The Figure 6 shows that the first three most-used terms in each subject are as follows: Topic 1(building, consumption, environment); topic 2 (design, water, decision); topic 3 (building, climate, fuel); topic 4 (decision, agriculture, improve); topic 5 (IoT, devices, consumption); topic 6 (urban, technology, industrial); topic 7 (engineering, efficiency, students); topic 8 (optimization, efficient, building).

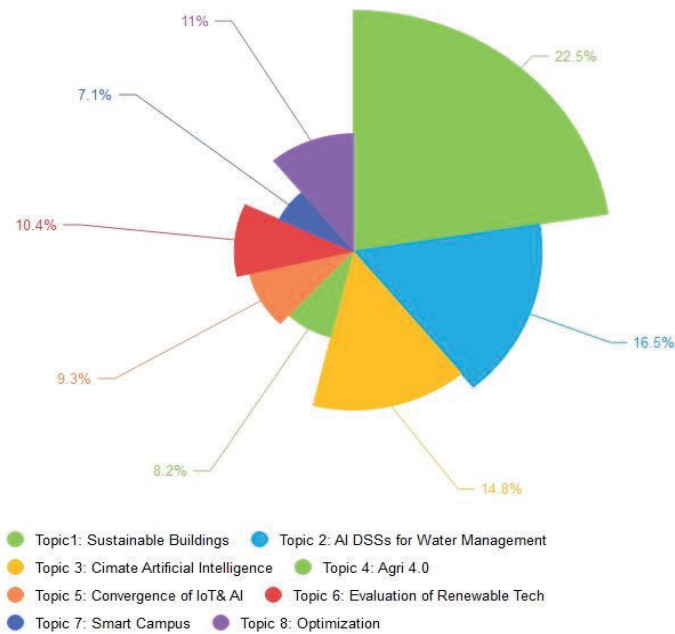

*Figure 4 The distribution of documents across topics*

*2.3. The evolution of topics over time*

Once we scoured the corpus for hidden topics, we determined how often they appear throughout time. Figure 5 depicts the ratios of all the eight topics (beginning in 2004 and extending into 2021). Since 2018 forward, topics have garnered a substantial amount of academic interest. Specifically, the first topic, which is about the design of sustainable buildings and minimizing energy usage via the application of artificial intelligence. This subject gained considerable attention between 2012 and 2014, but then slipped off the spotlight between 2015 and 2018. The discussions about AI-based evaluation of renewable energy solutions peaked around 2008 but then became less prominent until 2019. Climate artificial intelligence experienced two distinct phases, with the second one peaking in 2015 and 2016 and the first between 2009 and 2012; however, topic reached its apex in 2019 and 2020. The topic of AI for energy efficiency has shown a reasonably steady increase from 2013, with its greatest growth occurring between 2020 and 2021. In 2020, significant academic focus was given to AI-based DSSs for urban water management.

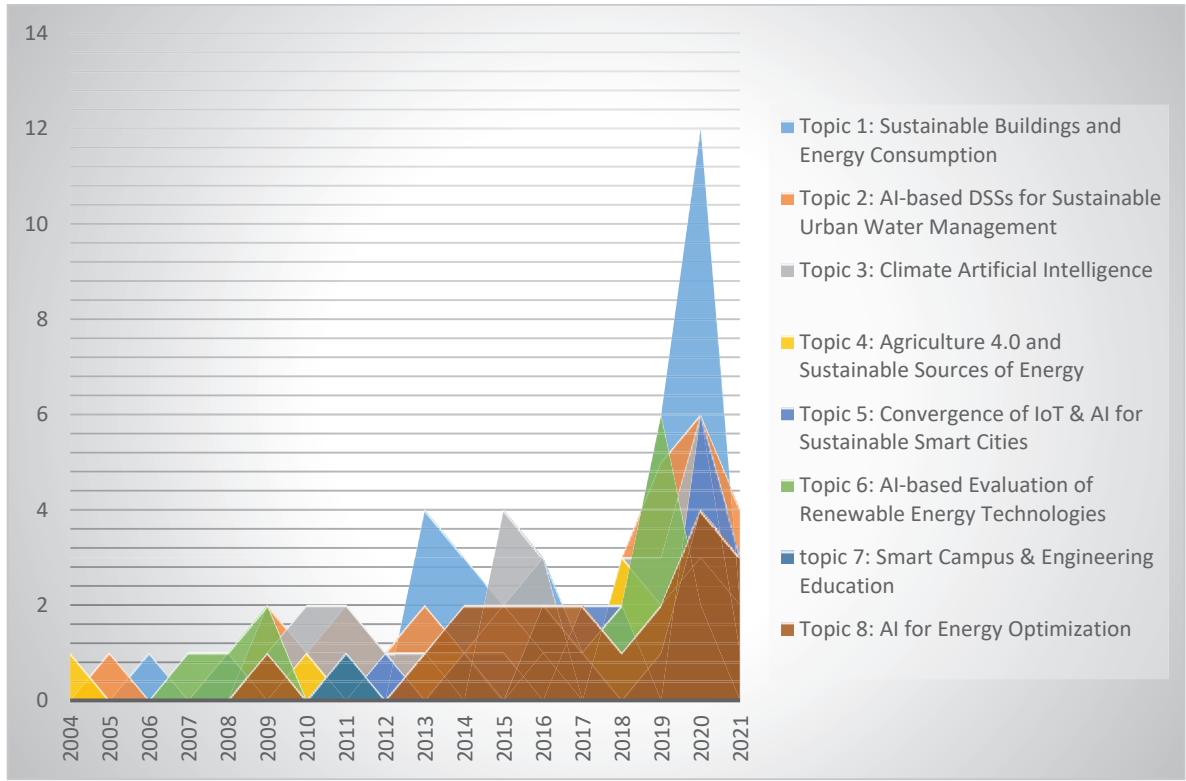

*Figure 5 The evolution of topics over time*

### 3.3. Content analysis to detect topics, sub-themes and cross-topic common themes

In this part of the paper, we conducted content analysis of detected topics for three purposes: First, to detect the general topics from articles; second, to identify the sub-themes from each topic, and third to find the cross-topic common themes.

*Topic 1: Sustainable Buildings and Energy Consumption*

The primary concerns of topic 1 are related to the design of automated and intelligent systems and the incorporation of cutting-edge technologies, particularly IoT and AI-based DSSs, in order to construct sustainable buildings. These buildings will be part of the sustainable cities initiative, which aims to promote sustainable energy consumption and smart grids.

One of the primary scholarly interests is the creation of sustainable buildings and smart grids for the purpose of reducing energy consumption. One way to accomplish this aim is to redefine the design and architecture of buildings, whether residential, public, commercial, industrial, or manufacturing. According to studies, the application of automation and intelligent systems in the construction of sustainable buildings will result in sustainable energy usage [32,33]. Several AI-based approaches are proposed to achieve a more sustainable building, including building management systems, knowledge-based engineering (KBE), fuzzy logic, neural networks, genetic algorithms, and Monte-Carlo simulation [34]. From a broad standpoint, sustainable building development falls under the umbrella of sustainable smart cities and reducing building energy consumption [35]. Additionally, scholars have drawn inspiration from nature and advocated regenerative design influenced by nature for pattern detection, prediction, optimization, and planning of buildings [36]. Additionally, scholars discuss the potential of AI in reducing CO2 emissions in buildings, suggesting that AI may be used to construct smart multi-energy systems, such as those found in industrial districts, resulting in significant energy savings and CO2 emission reductions (Simeoni, Nardin and Ciotti, 2018 ). As a result, sustainable building design would be a way to combat climate change.

Several additional studies integrate AI solutions with other cutting-edge technologies, most notably the Internet of Things and big data, to improve not only the design and optimization of sustainable buildings, but also the efficiency of their power usage (Chui, Lytras and Visvizi, 2018). For instance, one project focused on the application of IoT in public buildings in order to discover and anticipate energy usage trends [39]. A preceding study, for illustration, outlines the obstacles involved in understanding the semantics of IoT devices using machine learning models. Image Encoded Time Series has been identified as an alternate method to other statistical feature-based inference[35]. Sustainability analysts from [40] and [41] studies have also advocated for continual monitoring of sustainability metrics by integrating AI with DSSs or ambient intelligence.

Both residential buildings and plants and commercial buildings and offices have the same issue in regard to energy usage. Previous studies incorporated multi-objective and multi-attribute decision making modeling as well as impact evaluation of the emission outputs to help designers and manufacturers to make environmentally sustainable decisions about the designs and production of facilities [42]. Researchers also believe that in order to provide bulk energy consumption forecast, control, and management, simulation techniques could be utilized [15], for instance in public buildings, offices and factories. Due to new modes of consumption and distributed intelligence, the electrical power grids have been also influenced, and as a result, smart energy grids have been generated to achieve sustainability [43].

*Topic 2: AI-based DSSs for Sustainable Urban Water Management*

The second topic is sustainable water management, which includes utilizing AI to create DSSs for consumption and water usage. Forecasting, real-time monitoring, and customized and adjustable pricing and tariffs are the primary strategies. AI is used with other sophisticated technologies to assist in the development of a smart city.

The previous studies have postulated several approaches, such as optimization and AI-based decision support systems, for water infrastructure management [44], better delivery of public services of smart cities such as water treatment and supply [45], AI-based water pricing and tariff options [46] and sustainable water consumption [47]. For this goal, AI is integrated with recent technological advances in urban life. This includes using open source data, employing deep learning algorithms, and developing smart street lighting systems. Such decisions about social impacts of smartphone applications or smart travel behavior are also examined [48].

AI techniques are utilized to anticipate water resource management [49], such as water quality by adopting algorithms such as neuro-fuzzy inference system [50]. Real-time optimization of water resources and cloud technologies are integrated with visual recognition techniques and created to improve efficiency with irrigation systems [51]. A study conducted on ecological water governance implementation using AI found that including algorithms into the system yields higher-quality information and better prediction models for accurate evaluation of water quality [52]. AI may be used for tracking water use and demand as well as forecasting water quality, but it can also be used for estimating water infrastructure maintenance, monitoring dam conditions, water-related diseases and disasters [53] and water reuse [54].

By critiquing conventional decision support systems, research offer alternatives based on artificial intelligence, such as a systematic decision process [55], sustainability ranking framework based on Mamdani Fuzzy Logic Inference Systems to develop a sustainable desalination plant [56] or an comprehensive and flexible decision-making process fueled by social learning and engagement aimed at ensuring the urban water system's environmental and energy sustainability [57]. One research offers a unique DSS for analyzing the energy effect of each of the urban water cycle's macro-sectors, including assessing the system's energy balance and proposing potential energy-efficient solutions ( Puleo *et al.*, 2016).

*Topic 3: Climate Artificial Intelligence (Climate Informatics)*

Climate informatics, specially climate artificial intelligence as a new field of study is concerned with issues such as AI-based DSSs to reduce greenhouse gas emissions, optimizing grid assets, enhancing climate resiliency and reliability, increasing energy efficiency, forecasting energy consumption and modeling earth systems. Moreover, within this topic, scholars have addressed the issue of explainable and trustworthy AL models due to the controversial nature of climate change.

Climate change has compelled societies to seek alternate energy sources and fuels [59]. Climate informatics [60], such as several AI-based solutions, including novel algorithms and DSSs, have been hugely beneficial in lowering greenhouse gas emissions in the energy sector. By improving grid assets, and strengthening climate adaptability these innovations have greatly contributed to this ultimate goal [15]. Reliable and explainable artificial intelligence models, as advocated in prior studies, might help stakeholders and decision-makers achieve climate-resilient and sustainable development goals [61]. By integrating advanced machine learing techniques, AI can propose fresh insights in complex climate simulations in the field of climate modeling [62]. Energy consumption patterns might undergo considerable changes due to climatic change, which means AI forecasts can aid in estimating future energy use for various climate scenarios [63]. It's not only businesses and other organizations that are using AI algorithms these days—AI algorithms are also being utilized to foster sustainable urban growth and mitigate climate change by examining how future urban expansion will affect material and energy flows [64]. Fossil fuel, used as the primary energy source, is the primary contributor to human greenhouse gases that influence the climate. AI is extensively utilized for decreasing carbon footprints and for avoiding fossil fuel combustion [65] as prior studies show that AI can act as an automated carbon tracker [66]. Artificial intelligence-powered technologies may help investors in analyzing a company's climate effect while making investment choices [67]. By drawing attention to climate change through visualization techniques, they help to educate the public on the effects of climate change [68]

Ultimately, AI algorithms may provide great resources for climate change conflicts, including in the field of modeling earth systems [69], teleconnections [70], weather forecasting ( McGovern and Elmore, 2017), future climate scenarios [72], climate impacts [73] and climate extremes[74].

*Topic 4: Agriculture 4.0 and Sustainable Sources of Energy*

The fourth area that academics in the field of sustainable AI for energy extensively address is the development of smart agriculture and sustainable energy sources. The primary issue in this subject is how to combine advanced technologies like IoT, drones, and renewable energy with AI in order to create automated and real-time systems.

According to some researchers, the agriculture industry is suffering from an insufficient application of responsible innovation[75]. As a result, the researchers are calling for a system referred to as Responsible Agriculture 4.0, which incorporates drones, IoT, robotics, vertical farms, AI, and solar and wind power linked to microgrids [76–78]. When it comes to the productivity of agriculture, factors such as the cost of energy for cultivation are equally significant [79]. Based on the premise that most agricultural machinery operates on fossil fuels, it may potentially contribute to climate change. Thus, new energy solutions, and AI-based approaches are provided. One way in which bioproduction and renewable energy may positively influence sustainable agriculture and farming is via the development of bioproduction and renewable energy [80]. Proposing new AI methods to forecast agricultural energy use has also been researched [79]. biomass may also be used to provide sustainable energy in agriculture, and care should be taken to avoid any injuries [81]. Real-time alerting systems, AI-based DSSs, real-time DSS forecasting models, and alternative energy sources such as solar and wind play a vital role in sustainable agriculture [82]. Maximizing agricultural production and economic stabilization while minimizing the use of natural resources and their harmful environmental consequences may be accomplished using renewable energy and AI [82]. Artificial intelligence enables academics to provide accurate forecasts of agricultural energy use [83]. Especially, a drastic shift toward sustainability in agricultural practices has occurred because of its confluence with other cutting-edge technology, including sensors, DSSs, greenhouse monitoring, intelligent farm equipment, and drone-based crop imaging. [84].

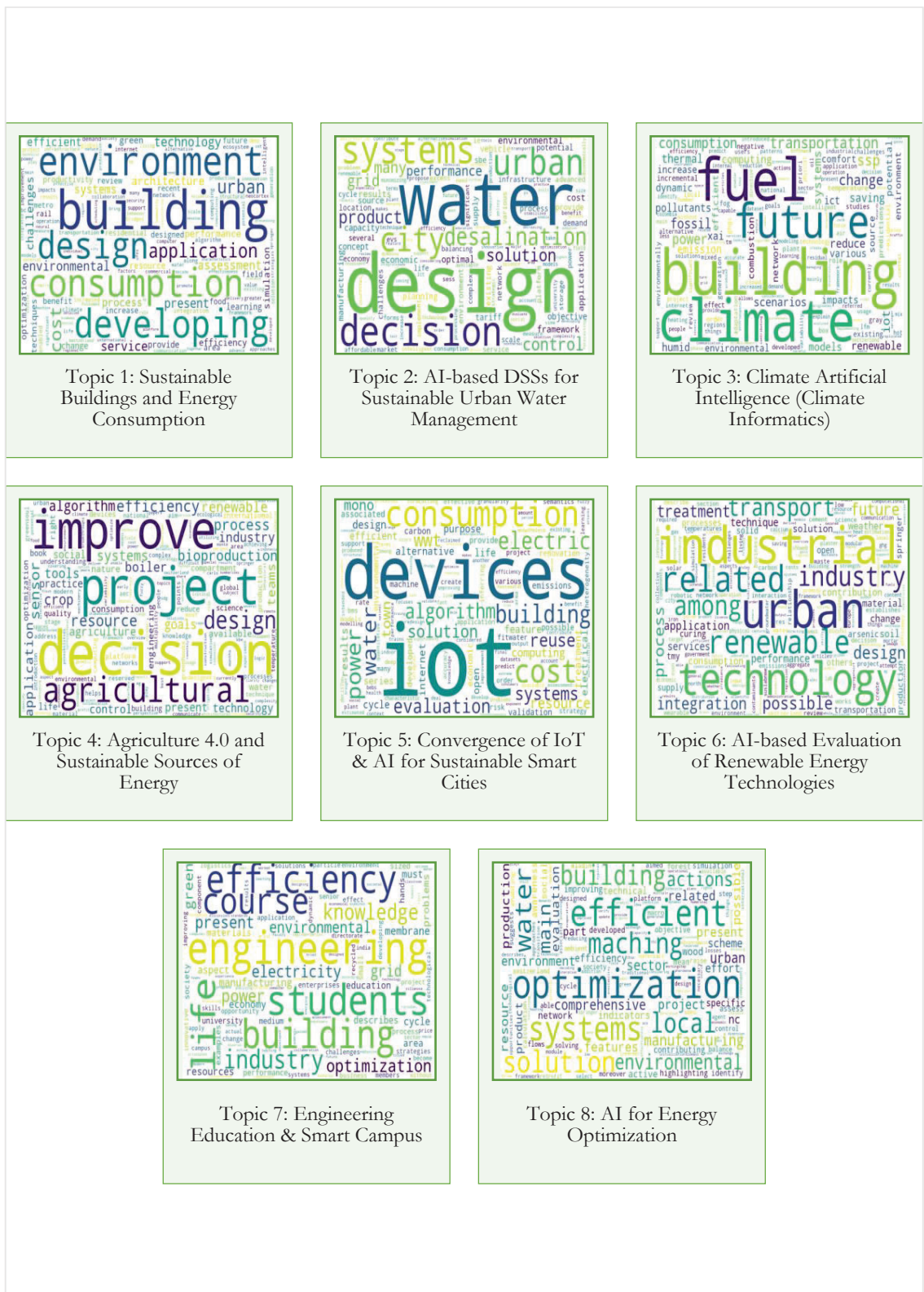

Figure 6 Topics detected by the combination of LDA+BERT+Clustering algorithms on sustainable AI in energy sector

*Topic 5: Convergence of IoT & AI for Sustainable Smart Cities*

A significant step in the implementation of sustainable energy solutions is to implement smart cities and services using internet of things technology. This topic exhibits how AI and IoT operate together to drive environmental progress. Much of this topic focuses on measure such as smart buildings, smart grid systems, green IoT, and smart campuses.

AI is used in tandem with a number of cutting-edge technologies for sustainable energy development, such as improved energy conservation [85] and building intelligent energy management [86] such as building management systems [35]. Internet of Things (IoT) is one of the most promising and pervasive technologies [85]; whose integration with AI has generated a revolution in the energy sector. There are many functions in creating sustainable energy in the IoT-enabled smart city dubbed City 4.0 [87] such as simulation and optimization of power plant energy sustainability [86]. City systems such as water and electricity, as well as other infrastructures, such as data analytics, will be driven by sensor and data collection in the smart city [87].

A significant use of IoT is in the design of intelligent buildings, which with AI included may support a goal of energy or water conservation [39,88], for instance, by educating the citizens on how to use energy more effectively and giving them warnings if they are using excessive amounts of energy. [89]. IoT is integral to modern grid development as well. In particular, it seeks to transform the traditional, fossil-fuel-based power grids with distributed energy resources and integrate it with cutting-edge technology such as artificial intelligence for improved grid management [90]. In the same manner, Blockchain has also been considered to be a viable alternative for smart cities. Fusing blockchain with AI may be leveraged for smart services, including energy load forecasting, categorizing customers, and evaluating energy load [91]. Smart connected devices such as IoT devices have successfully employed blockchain in time to retain these devices safe and secure in a blockchain network [92].

The effect of IoT and AI on agriculture and food sectors is also substantial [93,94]. Manufacturing facilities such as food factories and plants may be transformed more intelligent and more environmentally friendly via the use of IoT and AI, which merge with nonthermal and advanced thermal technologies [94]. Sustainable and green IoT are other topics covered in this subject. The two main objectives of the literature on green IoT are to increase the recyclability and usefulness of IoT devices, as well as to minimize the carbon footprints of such devices. The second objective is to incorporate more effective life cycle assessment (LCA) methods integrating artificial intelligence (AI) in order to cut costs and time [95]. Another of the many topics that apply to IoT is with developing smart campuses, which are carbon neutral, energy efficient, use less water, and are laced with various high-quality green energy tools [96] and smart teaching and learning platforms [97]. Researchers have identified the positive traits of IoT devices, but they've also forewarned about the possible risks of the devices and proposed various techniques for detecting weaknesses [93] or challenges regarding the heterogeneity of smart devices and their associated meta-data [35].

*Topic 6: AI-based Evaluation of Renewable Energy Technologies*

Scholarly interest has been generated by the discussion of leveraging AI for DSSs to enhance the efficiency of conventional system evaluations for renewable energy technologies. To a great extent, a sustainable future will depend on maximizing the use of energy sources that cannot be depleted [98]. Artificial intelligence is important for the survival of the future by leveraging a wide range of renewable energy technologies such as biomass energy, wind energy, solar energy, geothermal energy, hydro energy, marine energy, bioenergy, hydrogen energy, and hybrid energy [99]. AI is used to evaluate renewable energy solutions based on their cost of energy production, carbon footprint, affordability of renewable resources, and energy conversion efficiency [100]. Artificial intelligence will ensure the most effective use of these resources while also pushing for improved management and distribution systems [14]. Distributed energy management, generating, forecasting, grid health monitoring, and fault detection are also made more efficient by using automated AI systems [101]. AI can help disperse the supply and demand of energy in real-time and improve energy consumption and storage allocation (Sun, Dong and Liang, 2016).

To mitigate against the barrier of utilizing renewable energy technology, the following measures are taken: Renewable energy sustainability is evaluated [103]; in addition, the turbulent and sporadic character of renewable energy data is addressed [104]. One research group claims that standard techniques such as LCA and EIA (Environmental Impact Assessment) may be improved by developing more advanced digital intelligent decision-making systems, or DSSs. It is feasible that improved assessments of renewable energy sources may be achieved via intelligent and automated technologies [105]. With the smart mechanisms in place, long-term detrimental consequences can be calculated, as well as visible and invisible factors [106]. Artificial intelligence (AI) increases the adaptability of power systems, providing DSSs for energy storage applications [107]. For instance, to ensure more use of battery-electric buses, and minimize the effect on the power grids, the researchers developed an AI-powered DSS [108]. Another research leveraged AI to create a DSS for forecasting future energy consumption patterns, and to provide a solution for utilizing renewable energy alternatives [109].

*Topic 7: Smart Campus & Engineering Education*

It is possible to break down the discussions inside this topic into two distinct types: those about engineering education and those which deal with using AI and IoT to construct intelligent campuses to help maintain sustainability objectives. The two themes represent two elements of education: one dealing with the learning contents, and the other with behavioral outcomes of developing smart campuses. To build a model of smart campuses, we should focus on incorporating IoT into the infrastructure, with subsequent implementations of smart apps and services, with smart educational tools and pedagogies and smart analysis as well [97]. A smart campus is in charge of energy consumption scheduling, while its telecommunications infrastructure serves as the place where data transfers are conducted [110]. Integrating cutting-edge technology, a smart campus captures real-time data on energy usage, renewable energy power generation, air quality, and more [111].

Another point of view is that higher education should equip itself with relevant skills and competences to help in realizing long-term sustainable objectives [112]. The energy sustainability in this respect may be addressed via engineering education and engineering assistance for high-level strategic decision-making [113]. This objective can be achieved by using innovative instructional programs, alongside cutting-edge technology such as artificial intelligence and the Internet of Things. A living lab campus equipped with technology, as well as a deep well of talent and competency, may serve as a digital platform for education and sustainable growth [114]. For illustration, to support ongoing research, teaching, and learning on sustainable development, the University of British Columbia (UBC) implemented the Campus as a Living Laboratory project, which included AI and IoT and other cutting-edge technologies [115]. Furthermore, there have been several research done to help AI seamlessly integrate with current educational institutions in order to aid in sustainable development learning [116].

*Topic 8: AI for Energy Optimization*

Conventional optimization methods may be a roadblock for making progress toward sustainability, and AI-based solutions can help eliminate such roadblocks. Whilst renewable energy sources, like solar and wind, have many merits, there are some downsides to consider. They are usually not always available and often rely on the climate, which renders employing them complicated [117]. A proper optimization of energy may be utilized to minimize greenhouse gas emissions and cut energy usage. Efforts to reduce costs and side effects of energy consumption are facilitated using optimization models [118]. Computational and intelligent resources have enabled academics to progress with optimization problems by employing advanced AI methods. Manufacturers have developed numerous energy-efficient appliances for this reason. Even if the deployment of digital technologies in buildings will likely lead to improved energy efficiency, that is not the sole solution. Studies recommend implementing energy-saving measures that don't just target environmental variables, but also include building inhabitants' comfort and preferences, which is achievable via the integration of AI-augmented algorithms [119]. For illustration, AI algorithms that not only monitor current actions but also give real-time alerts and warnings to users and providers allow optimization to be significantly accelerated. Some approaches, such as algorithms that use energy consumption data to lower energy costs in buildings that use advanced AI, are only one example of how AI and advanced technology may be used to benefit society [120].

Weather has a direct effect on energy consumption, which is indisputable. To ensure the winter heating demand of non-residential buildings was calculated correctly, researchers used an optimized artificial neural network method to determine and forecast this need [121]. By utilizing AI along with the use of smart metering and non-intrusive load monitoring, one may improve energy efficiency by evaluating the electricity use of appliances [38]. Using a new approach, researchers found that the GP model was capable of making accurate predictions and a multi-objective genetic algorithm, NSGA-II, was also capable of optimizing sustainable building design [32]. The use of a fuzzy-enhanced energy system model to represent a route to a sustainable energy system has also been presented in another research [122]. The views of other researchers in the field include techniques based on artificial neural networks, evolutionary algorithms, swarm intelligence, and their hybrids, all of which rely on biological inspiration. These findings imply that sustainable energy development is computationally challenging conventional optimization, demanding advanced techniques [123].

## 4. Discussion, Theoretical Gaps, and Future Strands of Research

To identify the relevant research topics in the literature on artificial intelligence for sustainability in the energy industry, we performed a contextual topic modeling combined with qualitative cluster analysis. We went beyond previous approaches in developing this novel analysis by combining three algorithms of topic modeling (LDA, BERT, and clustering) with content analysis. In this research, eight academic topics were discovered including sustainable buildings and energy consumption, AI based DSSs for sustainable urban water management, climate artificial intelligence, agriculture 4.0 and sustainable sources of energy, convergence of IoT and AI for sustainable smart cities, AI-based evaluation of renewable energy technologies, smart campus and engineering education and AI for energy optimization. Concerns and problems addressed in each topic are summarized in Figure 7. The Figure illustrates that each topic addresses a number of specific issues, which some of them overlap.

For topic 1, the key problems are the importance of sustainable buildings for smart city development and smart grid services. The issue of AI and its application in decision-making, pricing, forecasting, and sustainable consumption are all addressed in this topic. To reach sustainability, various cutting-edge technologies are tied to AI. One problem which may be especially neglected is the use of AI technology to make buildings eco-friendlier and enhance their inhabitants' feeling of accountability toward sustainability. One approach might be to design real-time warning systems to ensure people are prohibited from excessive energy use, while also ensuring that they benefit from the AI-based solutions. Convergence research may also explore how green architecture is uniquely enabled to deal with complex issues, including environmental efficiency, such as using eco-lighting, natural ventilation, shading, green roofs, and artificial intelligence. Most of prior research focuses on eco-design and overlooks other factors of green architecture.

Topic 2 addresses sustainable urban water management via the use of AI-based DSSs. Conventional DSSs were under criticism from academics who suggested alternatives, and innovative approaches to DSSs were revealed, particularly with regard to water utilities in a smart city. The second discussion point, focused on sustainable consumption and real-time and predictive modeling, is also addressed in topic 2. Mitigating urban problems, notably air pollution, waste management, and wastewater management, are applicable here to exemplify how smart energy management leveraging AI improves environmental sustainability. Topic 3 deals with the connection between climate change and artificial intelligence, and the emergence of the climate informatics field. This topic highlights the role of trustworthy of explainable AI algorithms, an issue which is marginalized in other topics. As a result, a future potential study direction may be the development of ethical artificial intelligence in other topics to help with the sustainable management of energy. One prospective future study area is the confluence of smart grids, renewable energy, and 5G technology, since these technologies have the potential to generate enormous volumes of big data. Furthermore, the use of AI in transportation seems worthy of analysis, for example, with regard to traffic predictions, public transit planning, and so on.

The agricultural 4.0 and sustainable energy sources are examined in Topic 4. Many problems relevant to the subject of "prosperity, sustainable consumption, forecasting, and convergence with other automated and real-time technologies" are covered in this topic. There is only a limited body of studies dedicated to precision farming and digital mapping, but both developments promise to lead to better knowledge of the environment and to improved energy management. Precision farming by assessing soil nutrients, detecting humidity in the air, and monitoring crops allows farmers to leverage digital maps for better energy management and fight against climate change. Other related areas of study include developing automated working environments. It is worthwhile to investigate the effect that artificial intelligence and other green technologies will have on the working conditions of farmers and farm operators, since AI may help with deeper speculations of working conditions in farms.

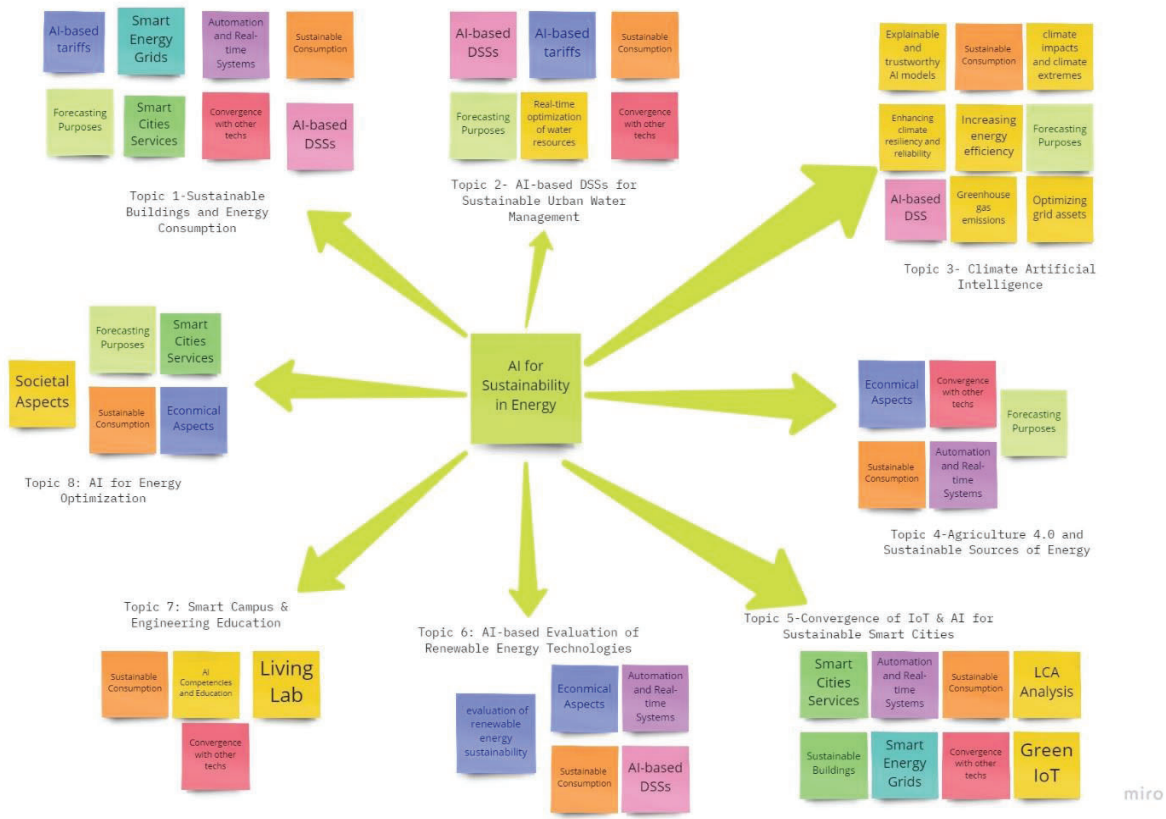

*Figure 5 Sub-themes extracted from each topic*

In Topic 5, convergent IoT and AI technologies for smart city development were addressed. The primary goal of this topic was to discuss issues around sustainable consumption, LCA analysis, and the development of intelligent energy grids. Pervasive Wi-Fi connection, due to its ability to save energy, is critical in this subject. Additionally, a significant problem is open data sharing in energy management. AI-based assessment of renewable energy technologies, such as DSSs, financial problems, sustainable consumption, and automated and real-time systems are all issues in this topic that focus on renewable energy. One potential study path in this topic involves the challenges that AI algorithms and models face when attempting to evaluate renewable energy solutions. Other sophisticated AI systems, such as deep learning, make use of supervised learning using human-annotated data, and thus they are limited when it comes to complicated situations.

The subject of smart campus and engineering education is examined in the seventh topic. Labs that facilitate continuous innovation are discussed in this article, as well as the idea of sustainable consumption, AI skills, and convergence with other technologies. There is an imperative requirement for further research to clarify how AI might be leveraged for practical learning and training for a range of stakeholders across businesses, farmers, residents, and employees in relation to energy management. AI is discussed in relation to energy optimization in Topic 8 of the study. This subject covers many elements of sustainable optimization, including forecasting, consumption, affordable pricing, and societal and financial impacts. However, there is a dearth of distributed energy resource optimization models, particularly due to the emergence of blockchain.

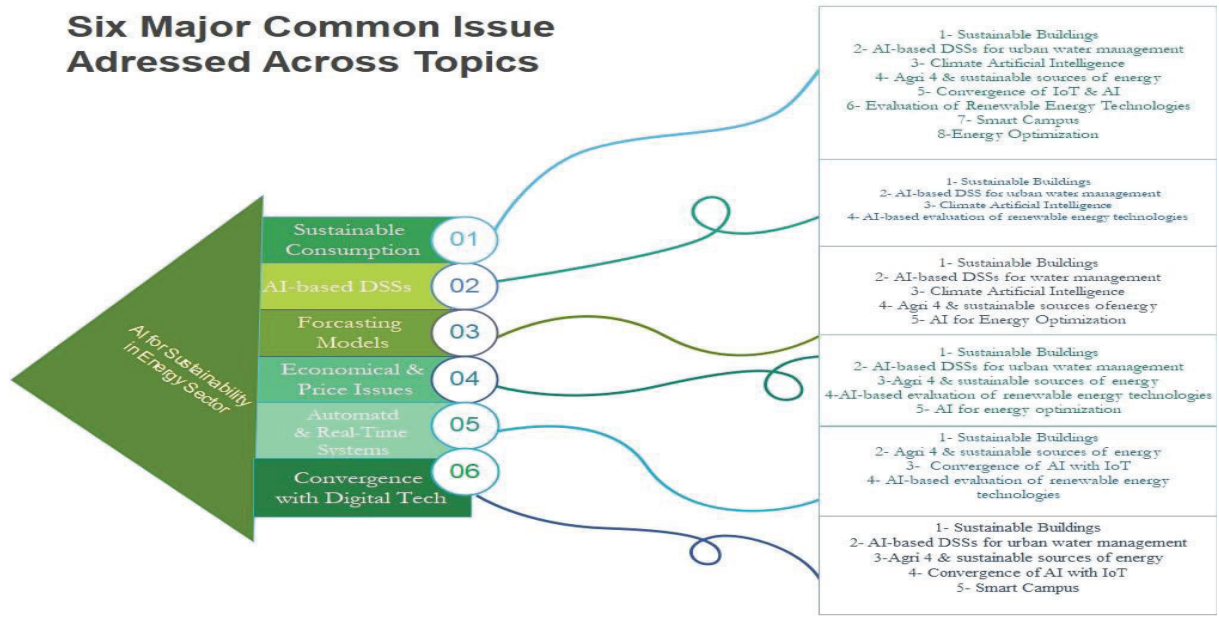

*Figure 6 Identified cross-topic common themes*

As shown in Figure 8, we discovered six core problems that were prevalent throughout the majority of the topics. For example, tariff and price models based on artificial intelligence are prevalent in topics 1 and 2; while economic issues in general are a concern in topics 4, 6, and 8. The dilemma of sustainable consumption is prevalent in all of these topics, demonstrating the critical role of AI in attaining sustainable energy use. Forecasting is inextricably connected to sustainable consumption, since more than half of the topics cover both; demonstrating the progress of AI forecasting algorithms for sustainable consumption. Forecasting, on the other hand, is not restricted to anticipating consumption patterns.

The topic's second significant recurring theme is the development of AI-based DSSs. The majority of research have contested traditional DSSs and devised decision-making systems based on artificial intelligence. Sustainable building, urban water management, climate change, and renewable energy evaluation have all been substantially influenced by AI-based DSSs. Automated and real-time systems enabled by artificial intelligence are also discussed in relation to buildings, agriculture, the Internet of Things, and renewable energy technologies. Scholars have combined various digital technologies to promote sustainability in the energy sector via the management of buildings, water, agriculture, IoT, and smart campuses.

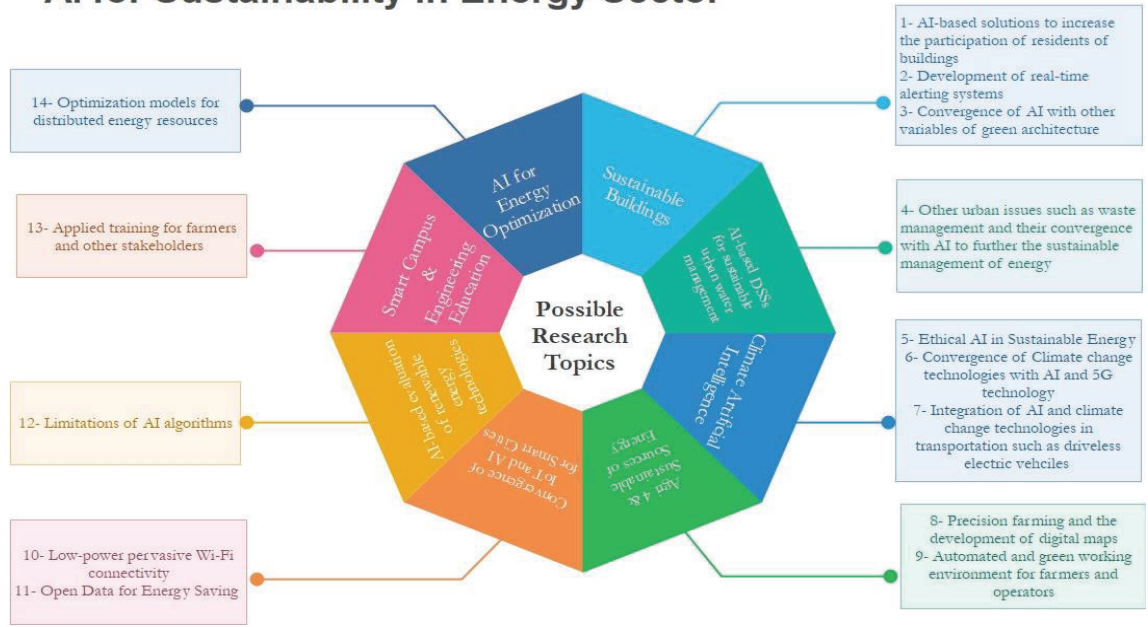

*Figure 7 Possible future streams of research pertaining to each topic*

## 5. Theoretical and Practical Contribution

*1.5. Theoretical Contribution*

Our results supplement existing work on sustainable AI and sustainable energy by delivering the following results. Results from this study provide and highlight a thematic map of the sustainable AI research topics existing in several fields, such as energy, ethics, and management. We developed a novel mixed-method approach, the contextual topic modeling and content analysis, to visualize the latent knowledge structures pertaining to AI and sustainability and energy. This yielded in a conceptual framework representing the main topics, subtopics and common terms in each topic pertaining to sustainable AI in energy. Using LDA and BERT, eight themes related to AI in the sustainability and energy sectors were discovered. We provided the most likely terms for each topic, as well as the distribution of articles and topics throughout time. Finally, by using a thematic analysis method, we identified and qualitatively analyzed the hidden themes.

Second, we examined and analyzed hidden sub-themes within each topic, as well as common themes between topics, using a content analysis method. Figure 8 illustrates the sub-domain themes within each topic, whereas Figure 9 depicts the common cross-topic themes. Our content analysis of each topic reveals six recurring themes: sustainable consumption, AI-based DSSs, forecasting models, economic and pricing problems, automated and real-time systems, and convergence with digital technology. To further our knowledge, we highlighted how these themes intersect across topics in order to articulate the commonalities across topics. These six separate but related topics demonstrate that sustainable AI solutions can be observed at a range of behavioral, decision-making, economic, operational, and technical dimensions. At the behavioral level, shifts in consumption patterns are illustrated; at the decision-making level, decision automation is outlined; at the economic level, personalized tariffing is demonstrated; at the operational level, automation and real-time operations are addressed; and at the technological level, convergence with other technologies is studied.

*2.5. Practical Implications*

This research provides energy engineers, social scientists, scientists, and policymakers with a variety of insights. Engineers may develop sustainable energy products and services. Energy scientists can also integrate sustainability considerations into their research and development of new energy sources such as renewable energy. In their discussions on AI and energy, social scientists may also emphasize ethical problems, including sustainability. Additionally, policymakers may create and construct new laws and policy initiatives aimed at mitigating the harmful effects of unsustainable energy on society and the environment.

## 6. Conclusion

To discover heavily discussed scholarly topics, our study utilized a new topic modeling technique. While this illustration depicts the trajectory of previous efforts, it also prompted us to propose a number of possible future research strands targeted at increasing energy sector sustainability via the application of artificial intelligence technology. The aim of this study is to further the conversation on sustainable AI and energy, as well as their intersection, in order to get a deeper understanding of how AI may be incorporated to achieve sustainability in the energy sector.

Scholz, M. Sustainability ranking of desalination plants using mamdani fuzzy logic inference systems. *Sustain.* **2020**, *12*, 631, doi:10.3390/su12020631.

57. Pearson, L.J.; Coggan, A.; Proctor, W.; Smith, T.F. A sustainable decision support framework for Urban water management. *Water Resour. Manag.* **2010**, *24*, 363–376, doi:10.1007/s11269-009-9450-1.

58. Puleo, V.; Notaro, V.; Freni, G.; La Loggia, G. Water and Energy Saving in Urban Water Systems: The ALADIN Project. In Proceedings of the Procedia Engineering; Elsevier Ltd, 2016; Vol. 162, pp. 396–402.

59. Afsordegan, A.; Sánchez, M.; Agell, N.; Zahedi, S.; Cremades, L. V. Decision making under uncertainty using a qualitative TOPSIS method for selecting sustainable energy alternatives. *Int. J. Environ. Sci. Technol.* **2016**, *13*, 1419–1432, doi:10.1007/s13762-016-0982-7.

60. Monteleoni, C.; Schmidt, G.A.; McQuade, S. Climate informatics: Accelerating discovering in climate science with machine learning. *Comput. Sci. Eng.* **2013**, *15*, 32–40, doi:10.1109/MCSE.2013.50.

61. Chakraborty, D.; Alam, A.; Chaudhuri, S.; Başağaoğlu, H.; Sulbaran, T.; Langar, S. Scenario-based prediction of climate change impacts on building cooling energy consumption with explainable artificial intelligence. *Appl. Energy* **2021**, *291*, 116807, doi:10.1016/j.apenergy.2021.116807.

62. Yuval, J.; O'Gorman, P.A. Stable machine-learning parameterization of subgrid processes for climate modeling at a range of resolutions. *Nat. Commun.* **2020**, *11*, 1–10, doi:10.1038/s41467-020-17142-3.

63. Fathi, S.; Srinivasan, R.; Ries, R. Campus Energy Use Prediction (CEUP) Using Artificial Intelligence (AI) to Study Climate Change Impacts. *Proc. 2019 Build. Simul. Conf. Rome, IBPSA, Italy* **2019**, doi:10.26868/25222708.2019.210874.

64. Blecic, I.; Cecchini, A.; Falk, M.; Marras, S.; Pyles, D.R.; Spano, D.; Trunfio, G.A. Towards a planning decision support system for low-carbon urban development. In Proceedings of the Lecture Notes in Computer Science (including subseries Lecture Notes in Artificial Intelligence and Lecture Notes in Bioinformatics); Springer, Berlin, Heidelberg, 2011; Vol. 6782 LNCS, pp. 423–438.

65. Kasim, H.; Hilme, H.; Al-Ghaili, A.; #2, R.O.; Hazim, M.; Muhmat, B.; #3, H.; Al-Ghaili, A.M. Future Fuels for Environmental Sustainability: Roles of Computing View project Developing a Digital Hub Framework in Inculcating Knowledge Sharing Practices for Malaysian Energy Sectors View project Future Fuels for Environmental Sustainability: Roles of Computing. *Int. J. Adv. Sci. Technol.* **2019**, *28*, 87–95.

66. Anthony, L.F.W.; Kanding, B.; Selvan, R. Carbontracker: Tracking and Predicting the Carbon